%% file: bare_jrnl_new_sample4.tex
\documentclass[lettersize,journal]{IEEEtran}
\usepackage{amsmath,amsfonts}
\usepackage{algorithm}
\usepackage{array}
\usepackage[caption=false,font=normalsize,labelfont=sf,textfont=sf]{subfig}
\usepackage{textcomp}
\usepackage{stfloats}
\usepackage{url}
\usepackage{verbatim}
\usepackage{graphicx}
\usepackage{cite}
\usepackage[dvipsnames]{xcolor}
\usepackage{booktabs}
\usepackage{multirow}
\usepackage{subfig}
\usepackage{subfloat}

\usepackage{algpseudocode}
\begin{document}
\bstctlcite{IEEEexample:BSTcontrol}

\title{Lifelong Learner: Discovering Versatile Neural Solvers for Vehicle Routing Problems}

\author{Shaodi~Feng, 
        Zhuoyi~Lin,
        Jianan~Zhou,
        Cong~Zhang,
        Jingwen Li,
        \\Kuan-Wen~Chen, ~Senthilnath Jayavelu,~\IEEEmembership{Senior Member,~IEEE},
       and~Yew-Soon~Ong,~\IEEEmembership{Fellow,~IEEE}
\thanks{Shaodi Feng and Kuan-Wen Chen are with the department of Computer Science at National Yang Ming Chiao Tung University (E-mail: fengshaodi411551029@nycu.edu.tw, kuanwen@cs.nycu.edu.tw)}
\thanks{Zhuoyi Lin (corresponding author) and J.Senthilnath are with the Institute for Infocomm Research, Agency for Science, Technology and Research (A*STAR), Singapore. (E-mail: zhuoyi.lin@outlook.com, J\_Senthilnath@i2r.a$-$star.edu.sg).
}
\thanks {Jianan Zhou is with the College of Computing and Data Science at Nanyang Technological University, Singapore. (E-mail: jianan004@e.ntu.edu.sg).}
\thanks {Cong Zhang is an independent researcher. (E-mail: cong.zhang92@gmail.com, cong030@e.ntu.edu.sg).}
\thanks{Jingwen Li is with the School of Computer Science, Sichuan Normal University, Chengdu, China. (E-mail: lijingwen@sicnu.edu.cn).}
\thanks{Yew-Soon Ong is with the School of Computer Science and Engineering,
Nanyang Technological University, Singapore, and also with the Center for Frontier AI Research, Agency for Science, Technology and Research
(A*STAR), Singapore. (E-mail: asysong@ntu.edu.sg).
}
}

\markboth{Journal of \LaTeX\ Class Files,~Vol.~14, No.~8, August~2021}%
{Shell \MakeLowercase{\textit{et al.}}: A Sample Article Using IEEEtran.cls for IEEE Journals}


\maketitle

\begin{abstract}

Deep learning has been extensively explored to solve vehicle routing problems (VRPs), which yields a range of data-driven neural solvers with promising outcomes. However, most neural solvers are trained to tackle VRP instances in a relatively monotonous context, e.g., simplifying VRPs by using Euclidean distance between nodes and adhering to a single problem size, which harms their off-the-shelf application in different scenarios. To enhance their versatility, this paper presents a novel lifelong learning framework that incrementally trains a neural solver to manage VRPs in distinct contexts. Specifically, we propose a lifelong learner (LL), exploiting a Transformer network as the backbone, to solve a series of VRPs. The inter-context self-attention mechanism is proposed within LL to transfer the knowledge obtained from solving preceding VRPs into the succeeding ones. On top of that, we develop a dynamic context scheduler (DCS), employing the cross-context experience replay to further facilitate LL looking back on the attained policies of solving preceding VRPs. Extensive results on synthetic and benchmark instances (problem sizes up to 18k) show that our LL is capable of discovering effective policies for tackling generic VRPs in varying contexts, which outperforms other neural solvers and achieves the best performance for most VRPs.

\end{abstract}

\begin{IEEEkeywords}
Vehicle Routing Problem, Learning to Optimize, Lifelong Learning, Transformer Architecture, Deep Reinforcement Learning.
\end{IEEEkeywords}

\section{Introduction}


\IEEEPARstart{I}{n} recent years, there has been a substantial increase in research on data-driven neural solvers for vehicle routing problems (VRPs), especially the travelling salesman problem (TSP) and capacitated vehicle routing problem (CVRP), emphasizing the centrality of VRPs as fundamental combinatorial optimization challenges in transportation and logistics ~\cite{cappart2023combinatorial,bengio2021machine, li2025unify,li2024multi}.
These solvers take advantage of various deep learning (DL) techniques to automatically discover efficacious VRP solving policies. However, most existing works focus on solving VRP instances in a relatively monotonous context. They often assume a random location of the customer node in terms of coordinates and simplify other problem aspects to be unchanged, such as Euclidean distance between nodes and the fixed problem size~\cite{kool2018attention,kwon2020pomo,wu2021learning,kim2021learning,bi2024learning}. While attaining promising results for the simplified VRPs, current neural solvers always encounter challenges when applied to real-world scenarios that feature more diverse contexts.

Several studies have aimed to enhance the neural solvers in their generalization performance across diverse problem sizes~\cite{kim2022scaleconditioned, chenrethinking} or distributions~\cite{zhang2022learning,jiang2022learning, gohshield}. Typically, these methods employ neural solvers originally developed for simplified VRPs, albeit adapting them to handle instances with varying contexts. They generate instances out of different problem sizes and distributions in batches or epochs for training. Nevertheless, to our knowledge, there is no technique proposed to ensure that the neural solvers stably transfer between problem sizes and distributions while avoiding catastrophic forgetting, i.e., forgetting the knowledge learned in preceding sources while adapting to new ones. Furthermore, current generalization techniques still fall short when applied to more generic VRPs, especially those with varying distance metrics, where even the same instance could have different optimal solutions owing to the distinct objective functions. Therefore, it is both practical and crucial to discover more versatile neural solvers for managing VRPs in broader contexts.


This paper presents a lifelong learning framework, including a lifelong learner (LL) with a dynamic context scheduler (DCS), to deliver versatile neural solvers that have the potential to tackle VRPs in distinct contexts (e.g., problem sizes, distributions, and distance metrics). Specifically, we employ a Transformer network as the LL, which is incrementally trained to construct solutions in a series of contexts. To this end, we propose an inter-context self-attention mechanism within LL to automatically transfer the knowledge learned from the preceding VRPs into the training process for the succeeding ones. Furthermore, we introduce the cross-context experience replay within DCS to preserve the learned policy for solving the preceding VRPs when training the LL to tackle a new VRP. The loss function is designed to optimize the policy for solving (current and preceding) VRP instances and regularize the self-attention for knowledge transfer. 
Compared to existing neural solvers, the proposed LL demonstrates the ability to continuously learn and adapt to new contexts as they emerge, without requiring the availability of data samples from all contexts. This not only reduces the time and computational costs associated with re-training models from scratch but also enables superior performance.
By exploiting the useful knowledge in the self-attention and replaying VRP instances in the preceding contexts, it is able to explicitly avoid the catastrophic forgetting of the knowledge acquired from previous VRPs.
We evaluate the lifelong learning method by training neural solvers in contexts of varying objective functions (i.e., distance metrics) and data sources (i.e., problem sizes), respectively. It is verified that the solvers are versatile enough to handle diverse contexts with favorable performance on both synthetic and benchmark instances, including those adapted from real-world ones with various node distributions. 
In summary, this paper's contributions are listed as follows: 
\begin{itemize}
    \item \textbf{Conceptual:} We introduce an early exploration of a lifelong learning framework aimed at training versatile neural solvers for more generic VRPs. This framework can effectively prevent catastrophic forgetting during incremental training for solving VRPs in distinct contexts.
    \item \textbf{Algorithmic:} Regarding the neural architecture, we propose an inter-context self-attention mechanism to maintain the useful knowledge from preceding VRPs in the process of training new VRPs. Meanwhile, the cross-context experience replay is further used to facilitate the LL to preserve the attained policy for solving the preceding VRPs when it is trained for new VRPs. 
    \item \textbf{Empirical:} The proposed method is evaluated over VRP instances with different contexts, including objectives and data sources. Extensive experiments on synthetic and benchmark instances show that LL consistently delivers strong performance across contexts and generalizes effectively to unseen metrics and problem sizes (up to 18k), validating its effectiveness as a generalist solver rather than solely excelling on specific contexts.

\end{itemize}

\section{Related works}
This section reviews neural solvers for VRPs and methods to enhance their generalization across varying distributions and sizes. We also outline lifelong learning approaches and highlight key applications in artificial intelligence.
\subsection{Neural Solvers for VRPs}
Recent neural solvers for VRPs could be divided into two categories: 
(1) \textit{Neural Construction Heuristics} construct the solution either sequentially or in a one-shot manner using learned heuristics without iterative adjustments. 
For example, the Pointer Network is introduced for solving TSP through supervised learning \cite{vinyals2015pointer}. Subsequent research extended its capabilities by training it with reinforcement learning for TSP \cite{bello2016neural} and CVRP \cite{nazari2018reinforcement}. 
The initial pivotal advancement arises from the utilization of the Transformer architecture \cite{vaswani2017attention}, whereby it is configured as the Attention Model~(AM) to address a spectrum of VRP variants~\cite{kool2018attention}.
To further improve the AM, POMO is proposed to enhance AM by leveraging solution symmetries \cite{kwon2020pomo}. It achieves state-of-the-art performance for solving TSP and CVRP among neural construction heuristics. 
(2) \textit{Neural Improvement Heuristics} acquire policies geared towards the incremental enhancement of an initially complete solution. Drawing inspiration from classical local search methodologies, several studies have ventured into the acquisition of learned procedures such as 2-opt, swap, and relocation operations, thus facilitating solution refinement \cite{wu2021learning,chen2019learning,lu2019learning,d2020learning,ma2021learning, wang2021bi}. Simultaneously, there have been endeavors to harness deep learning techniques to enhance the performance of established algorithms or solvers like Lin-Kernighan-Helsgaun (LKH)~\cite{kim2021learning}, Hybrid Genetic Search (HGS)~\cite{hottung2019neural}, and Large Neighborhood Search (LNS)~\cite{ xin2021neurolkh}. 
Despite effectiveness, current neural solvers are trained to tackle VRPs in a relatively monotonous context, without considering practical vehicle routing in more varying scenarios. Some approaches have been proposed to ameliorate the performance of neural solvers in different problem sizes or distributions~\cite{chenrethinking,kim2022scaleconditioned,zhang2022learning,jiang2022learning,bi2022learning,zhou2023towards, gohshield, luoboosting}. They typically train the neural solvers specialized for a single context, using varying instance sizes or distributions. However, they lack techniques to allow the neural network to be stably transferred, which could limit their performance owing to catastrophic forgetting. In this paper, we present an initial exploration of applying lifelong learning for solving VRPs in more generic contexts, e.g., the varying distance metrics that induce different objectives.


\subsection{Lifelong Learning}
Lifelong learning aims to prevent the catastrophic forgetting throughout the sequential training of a diverse set of tasks.
Approaches in lifelong learning generally can be grouped into three categories:
(1) \textit{Regularization-based methods}, such as LwF \cite{li2017learning} and EWC \cite{kirkpatrick2017overcoming}, estimate parameter or gradient importance from previous tasks and add regularization terms to the objective when learning a new task \cite{ zenke2017continual,aljundi2018memory};
(2) \textit{Memory-based methods} store and replay training samples of previous tasks \cite{lopez2017gradient} or store the gradient of previous tasks \cite{farajtabar2020orthogonal,lin2022trgp} to combat the forgetting of previous knowledge;
(3) \textit{Parameter isolation-based methods} prevent catastrophic forgetting by expanding 
the model architecture to accommodate learning a new task when needed \cite{rusu2016progressive,kang2022forget}.
In recent years, lifelong learning has been widely adopted for computer vision (CV) and natural language processing (NLP) tasks. 
For instance, in CV, lifelong learning enables models to sequentially learn from samples with varying class annotations, allowing accurate detection and classification of previously learned classes \cite{shmelkov2017incremental,zhao2022static}.
In NLP, recent work proposes to mitigate the loss of generality in handling diverse distributions and NLP tasks with lifelong learning during the fine-tuning \cite{razdaibiedina2023progressive,chen2023lifelong}. 
In this paper, we provide an early attempt to exploit lifelong learning to deliver more versatile neural solvers for VRPs. 


\section{Preliminaries}
In this section, we present the problem definition of VRPs, 
and outline the neural construction heuristics, which is one of the most prevalent paradigms for solving VRPs.

\subsection{VRP Definition} 
A VRP instance can be defined over a graph $\mathcal{G}$ with nodes $\mathcal{V}=\left\{v_0, v_1, \cdots, v_n\right\}$, where each node $v_i\!\in\! \mathcal{V}$ is featured by $o_{i} \in \mathcal{O}$. $\mathcal{E}=\{e_{ij}|v_i, v_j\in \mathcal{V}; v_i\neq v_j\}$ denotes edges featured by their weights $w_{ij}$.  The solution to a VRP instance is a tour $\pi=\left(\pi_{1}, \pi_{2}, \cdots, \pi_{T}\right)$, i.e., a node sequence of length $T$, with $\pi_{j}\!\in \! \mathcal{V}$. A solution is feasible if it satisfies problem-specific constraints. The objective of solving a VRP instance is to find the optimal tour $\pi^{*}$ with the minimum total weight.

\noindent\textbf{TSP.} The node feature in TSP is the coordinates $o_{i}=(x_i, y_i)$. The edge weight represents the distance between nodes, for example, $w_{ij}^{E}=\sqrt{(x_i-x_j)^2+(y_i-y_j)^2}$ in a Euclidean space.
A tour is feasible if it forms a closed loop, and each node is visited exactly once.

\noindent\textbf{CVRP.} The node feature in CVRP is represented as a 3-tuple $o_{i}=(x_i, y_i, d_i)$, where $(x_i,y_i)$ denotes the coordinates and $d_i$ denotes the demand. Specially, the demand of the depot node $v_0$ is 0, i.e., $d_0=0$.
The edge weight is the distance between any pair of nodes. A tour is feasible if (1) each node, except the depot, is visited exactly once; and (2) the total demand in each route (i.e., a subtour starting and ending at the depot) does not exceed the vehicle capacity $Q$.

\noindent\textbf{Distance Metrics.} Although current neural solvers primarily focus on Euclidean distances between nodes, real-world scenarios often necessitate a variety of distance metrics~\cite{ji2001planning,duman2004precedence}. The commonly used distance metrics beyond Euclidean distance are (1) \emph{Manhattan} distance: the sum of the absolute differences between $x$ and $y$ coordinates, i.e., $w_{ij}^{M}=|x_i-x_j|+|y_i-y_j|$; (2) \emph{Chebyshev} distance: the maximum of the absolute differences between $x$ and $y$ coordinates, i.e., $w_{ij}^{C}=\max{(|x_i-x_j|, |y_i-y_j|)}$. 

\noindent\textbf{Lifelong Setting.} To achieve a generalist neural solver toward solving more generic VRPs, we propose to formulate the VRP as a lifelong learning problem, which is split in a sequence of $\textit{M}$ contexts $\mathcal{M}_m$, $m\in\{1,…,M\}$. For each context (e.g., distance metric), instances $S \in \mathcal{S}_m$ are sampled from a normal distribution, which are further constructed as VRP graphs $\mathcal{G}$. The neural solver processes contexts sequentially, observing one context at a time.
Consequently, it is not feasible to optimize all observed contexts simultaneously. However, a small amount of instances can be stored in a limited memory and used for future rehearsal.

 \begin{figure*}[!t]
 \vspace{-0.1in}
    \centering
    \includegraphics[width=0.97\linewidth]{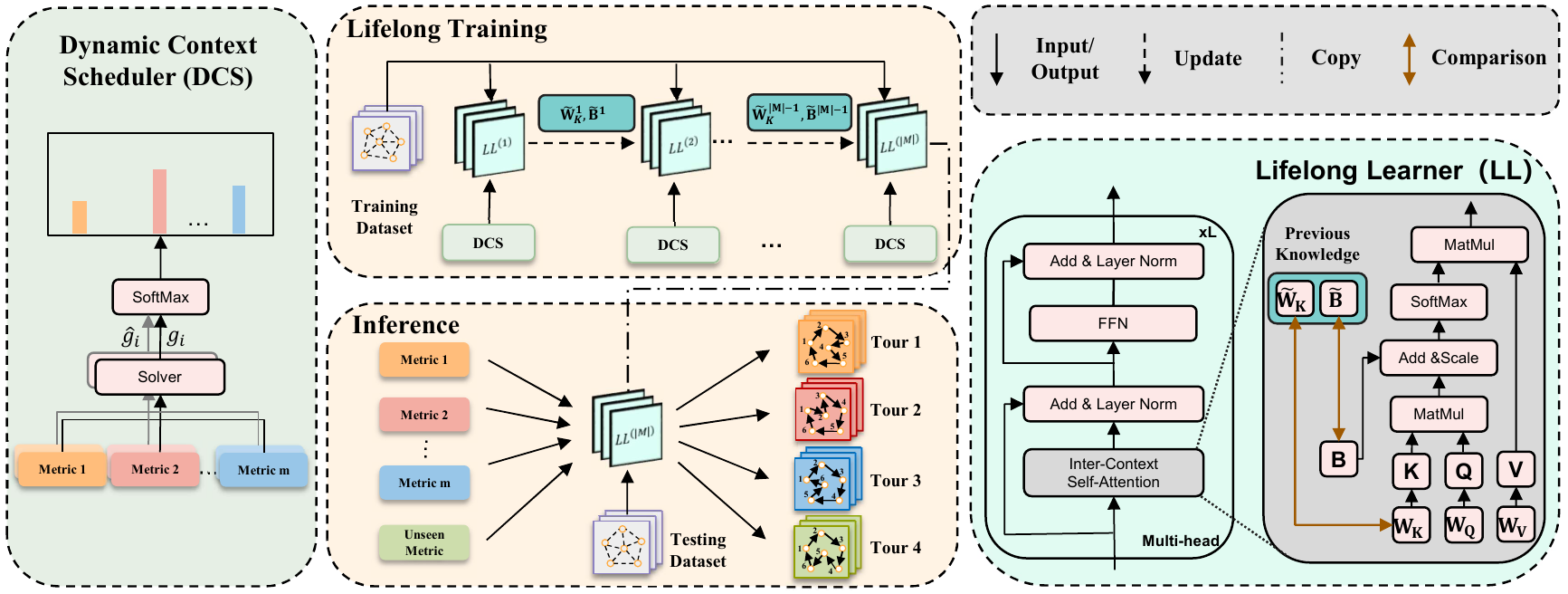}
        \vspace{-0.1in}
    \caption{An overview of the lifelong learning framework for VRPs. We present the lifelong training and inference process in the middle. The dynamic context scheduler (DCS) and lifelong learner (LL) are illustrated on the left and right sides, respectively.}    
    \label{fig: framework}
\end{figure*}

\subsection{Neural Construction Heuristics} \label{sec:NCH}

Learning construction heuristics is a prevalent paradigm for solving VRPs~\cite{bello2016neural,kool2018attention,kwon2020pomo}. Specifically, solutions are built by sequentially selecting valid nodes, a process that can be modeled as a Markov Decision Process (MDP). Given a state comprising the instance information and the current partial solution, the agent iteratively selects a valid node from the remaining ones until a feasible solution is constructed. The solution construction policy is typically parameterized by an LSTM or Transformer network $\theta$, which, at each step, infers a probability distribution over valid nodes. A node is then sampled and added to the partial solution. The probability of constructing a tour $\pi$ is factorized as $p_{\theta}(\pi|\mathcal{G}) = \prod_{t=1}^{T} p_{\theta} (\pi_{t}|\mathcal{G}, \pi_{<t})$, where $\pi_t$ and $\pi_{<t}$ denote the selected node and the current partial solution at step $t$, respectively.
The REINFORCE~\cite{williams1992simple}, one of the policy gradient DRL algorithms, is opted to update $\theta$:
\begin{equation}
    \label{eq:reinforce}
    \nabla_{\theta} \mathcal{L}(\theta|\mathcal{G}) = \mathbb{E}_{p_{\theta}(\pi|\mathcal{G})} [(c(\pi)-b(\mathcal{G})) \nabla\log p_{\theta}(\pi|\mathcal{G})],
\end{equation}
where $c(\pi)$ is the cost (e.g., total length) of the tour $\pi$, and $b(\cdot)$ is an action-independent baseline function to reduce the variance of the gradient estimation.


\section{Methodology}\label{method}
In this section, we present our lifelong learning framework to assist in training neural VRP solvers in diverse contexts. To mitigate the catastrophic forgetting issue, we first exploit a Transformer network as the LL, which is trained on VRPs with varying contexts (e.g., varying distance metrics or problem sizes). 
We propose an inter-context self-attention mechanism within the Transformer network \cite{vaswani2017attention} to enable automatic knowledge transfer between contexts, and a cross-context experience replay to regularly review and integrate knowledge from prior contexts, further supporting learning across contexts.
Without loss of generality, we implement the LL framework on POMO~\cite{kwon2020pomo}, where the self-attention encoder generates node representations and the decoder outputs node selection probabilities. Our framework is applicable to other Transformer-based heuristics~\cite{kool2018attention,kim2022symnco,huang2025rethinking,luo2024neural}. The following sections detail our LL framework across distance metrics.

\subsection{Overall Framework}
The overview of the proposed lifelong learning framework is illustrated in Figure \ref{fig: framework}, including a \emph{lifelong learner (LL)}
with a \emph{dynamic context scheduler (DCS)}.
Concretely, the LL is trained on a set of tasks, each characterized by a unique context (e.g., distance metric).
To facilitate effective lifelong training and alleviate the catastrophic forgetting issue, the distance metrics used in the previous tasks should be revisited when learning a new task. Therefore, we propose an inter-context self-attention mechanism within LL and a cross-context experience replay within DCS.
This framework enables efficient training with only a few distance metrics, but delivers a versatile LL (i.e., neural solver) capable of favorably solving VRPs with various distance metrics. 

\subsection{Lifelong Learner}
A LL is supposed to alleviate the catastrophic forgetting issue during training with varying contexts, i.e., the learned knowledge from the preceding metric is overwhelmed by the training for the current metric.
Meanwhile, different contexts may share a similar problem structure, so the learned policies from the preceding contexts could be used to facilitate the training in succeeding contexts. For example, TSPs or CVRPs with different distance metrics can be represented by similar graphs, featuring the same node attributes and the aim to minimize the total distance of the tour.
The only difference lies in the definition of distances between nodes.
To mitigate the catastrophic forgetting and leverage the common problem structure, we propose an inter-context self-attention mechanism that empowers POMO to automatically capture the cross-context knowledge, which could be elegantly transferred among contexts and benefit the lifelong training. 

Specifically, the inter-context self-attention employs a learnable pair of key matrix $\mathbf{{W}_K}$ and bias matrix $\mathbf{B}$
to implicitly preserve the information in the current context, which is transferred forward to guide the training of the LL in the following context. 
As shown in the right side of Figure \ref{fig: framework}, given the input $\mathbf{X}\in \mathbb{R}^{n\times d}$ in the current context, where $n$ and $d$ denote the number of nodes and hidden dimension, the LL first attains the query ($\mathbf{Q}$), key ($\mathbf{K}$) and value ($\mathbf{V}$) matrices: 
\begin{equation}
\mathbf{Q} = \mathbf{X}\mathbf{W}_Q \text{,  } \mathbf{K} = \mathbf{X}\mathbf{W}_K \text{, and } \mathbf{V} = \mathbf{X}\mathbf{W}_V,
\end{equation}
where $\mathbf{W_Q}$, $\mathbf{W_K}$, $\mathbf{W_V}\in \mathbb{R}^{d\times d}$ are learnable matrices in linear transformations. Then, the inter-context self-attention mechanism in our LL is defined below:
\begin{equation} 
\begin{aligned}
\text{Attention}(\mathbf{Q},\mathbf{K}, \mathbf{V}) = \text{Softmax}(\frac{\mathbf{Q}\mathbf{K}^{T}+\mathbf{B}}{\sqrt{d}})\mathbf{V},
\end{aligned}
\label{attention}
\end{equation}
where $\mathbf{B}\in \mathbb{R}^{n\times n}$ is the learnable bias matrix. 
Note that $\mathbf{W_K}$ and $\mathbf{B}$ in the current context $\mathcal{M}_m$ are initialized by their corresponding parameters learned from the previous context $\mathcal{M}_{m-1}$ (denoted as $\mathbf{\widetilde{W}_K}$ and $\widetilde{\mathbf{B}}$). Similarly, the trained $\mathbf{W_K}$ and $\mathbf{B}$ in $\mathcal{M}_{m}$ will serve as the previous knowledge for the next context $\mathcal{M}_{m+1}$.
By continuously consolidating $\mathbf{\widetilde{W}_K}$ and $\widetilde{\mathbf{B}}$ via the inter-context self-attention in each context, the previous knowledge can be effectively preserved to mitigate catastrophic forgetting of past contexts and meanwhile facilitate the training process for the current context. In addition, our inter-context self-attention mechanism is multi-headed by parallelly running multiple computations of Eq.~(\ref{attention}), so that each head learns attention weights and node embeddings within different subspaces:
\begin{equation} 
\scriptsize
\label{multi-head self-attention}
\begin{aligned}
\text{Multi-head}(\mathbf{Q}, \mathbf{K}, \mathbf{V}) &= \text{Concat}(\text{head}_{1}, \cdots, \text{head}_{h})\mathbf{W}^{O},\\
\text{head}_{i} &= \text{Attention}(\mathbf{X}\mathbf{W}_{Q}^{i}, \mathbf{X}\mathbf{W}_{K}^{i}, \mathbf{X}\mathbf{W}_{V}^{i}),
\end{aligned}
\end{equation}
where $h$ denotes the number of heads, and $\mathbf{W}_{Q}^{i}$, $\mathbf{W}_{K}^{i}$, $\mathbf{W}_{V}^{i}\in \mathbb{R}^{d\times d/h}$, $\mathbf{W}^{O}\in \mathbb{R}^{d\times d}$ are learnable matrices. The key matrices $\{\mathbf{\widetilde{W}_K}^i\}_{i=1}^h$ and bias matrices $\{\widetilde{\mathbf{B}}^i\}_{i=1}^h$ (in heads) are reshaped from $\mathbf{\widetilde{W}_K}$ and $\widetilde{\mathbf{B}}$, which are transferred across contexts. By doing so, the inter-context self-attention remains the same output dimension $\mathbb{R}^{n\times d}$ as the input.
We apply the proposed self-attention in each encoder layer, and keep other components unchanged, i.e., feed-forward subnetwork (FFN), skip-connection (Add), and layer normalization (Layer Norm), as shown in Figure \ref{fig: framework}.

It is noteworthy that different from most existing lifelong learning approaches, our LL is able to effectively handle different contexts in VRPs by preserving the previous knowledge in $\mathbf{\widetilde{W}_K}$ and $\widetilde{\mathbf{B}}$, e.g., distinct distance metrics or varying problem sizes, which accomplishes more versatile neural solvers for VRPs. 
This is the first time the lifelong learning is explored to train DRL based models of VRPs. In the next section,  we present a cross-context experience replay to further alleviate the catastrophic forgetting of DRL based models in our lifelong learning framework.

\subsection{Dynamic Context Scheduler}
Experience replay is a common technique in lifelong learning to ameliorate deep models against catastrophic forgetting. The main principle behind it is to balance the training effort in either learning from new experiences or revisiting old ones, with the aim to continuously improve the performance on both previous and new tasks~\cite{lopez2017gradient,chaudhry2019continual,farajtabar2020orthogonal,lin2022trgp}.
Most existing experience replay techniques review previous knowledge by training the current model with a few data samples from previous tasks, which distributionally deviate from the current task~\cite{wang2022continual, wang2023comprehensive}. 
In our lifelong learning for VRP tasks with different contexts, the distribution of node coordinates and demands could remain the same across tasks and the LL should more focus on learning a policy across different distance metrics that directly affect the optimization objective. 
Hence we propose to replay the previous distance metrics rather than revisiting training samples in past contexts. A simple yet effective solution is to generate a fixed number of training samples for each previous metric, which are used to train the LL for a few epochs before training on a new distance metric.

The above experience replay revisits each previous metric by the same number of samples, implicitly assuming that the knowledge in all previous contexts is equally important to the training process of the current context.
Further, we propose a dynamic context scheduler (DCS) to guide the cross-context experience replay during the lifelong training. 
For each epoch, we assess the performance (i.e., gap) of the LL in all contexts to measure their hardness: 
\begin{equation}
\label{eq:hardness}
g_i= \frac{1}{K} \sum_{k=1}^{K} \frac{c(\pi_k^i) - c(\Bar{\pi}_{k}^i)}{c(\Bar{\pi}_k^i)},
\end{equation}
where $K$ is the number of validation instances. $c(\pi_k^i)$ and $c(\Bar{\pi}_k^i)$ denote the cost of solutions to the $k_{th}$ instance defined with the $i_{th}$ distance metric, which are constructed by LL and an efficient solver\footnote{In this paper, we apply LKH3 \cite{helsgaun2017extension} to efficiently generate near-optimal solutions for VRPs. Note that we only need to run LKH3 once as the validation instances are fixed throughout the lifelong training process.}, respectively. Given Eq.~(\ref{eq:hardness}), we incentivize the LL to more frequently revisit metrics that demonstrate suboptimal performance. 
We also consider performance differences for each metric between the current and last epoch, enabling it to more thoroughly revisit knowledge that has been significantly forgotten.
The probability distribution of metrics is calculated by DCS as:
\begin{equation}
\label{eq:metric_prob}
    p(w^{i})= \frac{\exp(g_{i} / \eta) + \exp (g_{i} - \hat{g}_{i})}{\sum_{j=1}^{|J|}\exp(g_{j}/ \eta)+ \exp (g_{j}-\hat{g}_{j})},
\end{equation}
where $\eta$ is the temperature parameter to regulate the entropy of probabilities, from which a metric is drawn to generate instances in each epoch. $\hat{g}_{i}$ denotes the performance of $i_{th}$ metric in the last epoch.
Consequently, during each epoch, training instances for each metric are generated based on the estimated probability distributions $p$ and used to train the LL.

\subsection{Loss Function}
The loss function of our lifelong learning framework comprises two components. 
The first component aims to uphold the constancy of the attention map throughout contexts, thereby alleviating catastrophic forgetting.
Concretely, we calculate the disparity between the current parameters ($\mathbf{W_K}$, $\mathbf{B}$) and the implicit knowledge attained from the preceding context ($\mathbf{\widetilde{W}_K}$, $\widetilde{\mathbf{B}}$) by a weighted $L1$-norm:
\begin{equation}\label{eq:lifelongloss}
\scriptsize
\mathcal{L}_r=\left\|\nabla_{\mathbf{\widetilde{W}}_k} \mathcal{J} \odot(\mathbf{W_k}-\mathbf{\widetilde{W}}_k)\right\|_1+\left\|\nabla_{\widetilde{\mathbf{B}}} \mathcal{J} \odot(\mathbf{B}-\widetilde{\mathbf{B}})\right\|_1,
\end{equation}
where $\odot$ denotes the Hadamard product. $\nabla_{\mathbf{\widetilde{W}}_k}\mathcal{J}$ and 
$\nabla_{\widetilde{\mathbf{B}}} \mathcal{J}$ are gradients of the loss function regarding $\mathbf{\widetilde{W}}_k$ and $\widetilde{\mathbf{B}}$ in the previous context, respectively.
Intuitively, the significance of parameters is proportional to the magnitude of gradients in the training process, and thus more penalties might be assigned to the parameters with greater importance in Eq.~(\ref{eq:lifelongloss}). 
We also empirically observe that imposing penalties on alterations of attention maps, i.e., ($\mathbf{W_K}$, $\mathbf{B}$), facilitates the preservation of information from preceding contexts. 

Most neural construction solvers (e.g., POMO~\cite{kwon2020pomo}) apply the REINFORCE algorithm to estimate gradients of the expected reward.
Once a set of solutions $\{ {\tau}^1, {\tau}^2, ..., {\tau}^N \}$ are sampled for a given training instance $\mathcal{G}$, the reward of each solution ${\tau}^j$ is denoted as $R({\tau}^j)=-c(\tau^j)$, where $c(\cdot)$ calculates the total tour distance. To maximize the expected reward, we update the neural construction solver by the approximated gradient:
\begin{equation}\label{eq:reinforce1}
 \mathcal{L} \approx -\frac{1}{N} \sum_{j=1}^N\left(R\left(\boldsymbol{\tau}^j\right)-b^j(\mathcal{G})\right)  \log p_\theta\left(\boldsymbol{\tau}^j \mid \mathcal{G}\right),
\end{equation}
where $\theta$ is the parameters of the policy network, and $b^j(\mathcal{G})$ is a baseline function to reduce the variance of the sampled gradients. We adopt the shared baseline from POMO:
\begin{equation}
b^j(\mathcal{G})=b(\mathcal{G})=\frac{1}{N} \sum_{j=1}^N R(\boldsymbol{\tau}^{j}), \quad \forall j \in \{1, \ldots, N\}.
\end{equation}
Overall, we jointly optimize the LL with the loss function defined by $\mathcal{J} = \mathcal{L} + \alpha \mathcal{L}_r$,
where $\alpha$ is a hyper-parameter to control the strength of regularization.
The primary term $\mathcal{L}$ is employed to train the policy across various VRP contexts during the lifelong learning.
The regularization term $\mathcal{L}_r$ is used to enhance the attention mechanism, preventing catastrophic forgetting and facilitating effective knowledge transfer among contexts.
Note that we present the training algorithm of LL parameterized by POMO~\cite{kwon2020pomo}, where we regard metrics as contexts for an example. A similar training procedure can be easily applied to other neural VRP solvers, with different contexts (e.g., varying problem sizes).

\input{table/performance}
\section{Experiments}\label{exps}

In this section, we empirically evaluate the performance of our proposed lifelong learner (LL) to tackle VRPs in distinct contexts.
Firstly, we compare it with other baselines to demonstrate its effectiveness on two representative VRPs (i.e., TSP and CVRP). 
Then, we design ablation experiments to demonstrate the efficacy of each component in our lifelong learning framework.
To show the versatility of LL for tackling generic VRPs in varying contexts, we apply our approach to solve cross-size problem instances.
Finally, we conduct evaluation on unseen metrics and benchmark datasets to further demonstrate its generalization ability. 

\subsection{Experimental Settings}\label{es}
In our LL, cross-context experience replay is performed by generating an additional 20\% of training instances for each previous distance metric. This process can be further combined with a DCS module, resulting in the LL+DCS variant. As described, we train and test our methods on three exemplar distance metrics (Euclidean, Manhattan, and Chebyshev). Consistent with POMO \cite{kwon2020pomo}, we evaluate all methods on problem instances of size 100. Specifically, we use Adam \cite{kingma2014adam} as the optimizer with a learning rate of $10^{-4}$ and weight decay of $10^{-6}$. The number of attention heads $h$ and Transformer blocks $L$ are set to 8 and 6, respectively.
For a fair comparison, we train until convergence for each approach, and our observations matched the suggestions from \cite{kwon2020pomo}. That is, most of the training can be completed in 200 epochs.
In addition, we follow existing work \cite{kwon2020pomo} to generate instances for both TSP and CVRP.
The experimental setup encompasses instances with 100 nodes, employing two-dimensional Euclidean distance to compute inter-city distances, with the goal of minimizing the overall travel distance. The node coordinates originate from a uniform distribution spanning the range [0, 1] independently for both dimensions. 
For CVRP, the demands of each non-depot city are randomly sampled from the integers in the range [1, 9], and the capacity $Q$ of the vehicle varies with the problem size, where we set $Q^{20}=30$, and $Q^{40}=35$. When the problem size is larger than or equal to 50 ($n	\geq 50$), we set the capacity $Q=30+n/5$.
We will release the source code upon acceptance.

\smallskip\noindent\textbf{Baselines.}
We benchmark against several baselines, encompassing different optimization problems. Regarding TSP, we compare our methods with the following baselines:
(1) Concorde \cite{concorde2006}, an exact solver to get the objective values of the optimal solutions for TSP.
(2) LKH3 \cite{helsgaun2017extension}, a state-of-the-art heuristic solver employed to obtain benchmark solutions.
(3) OR-Tools \cite{ortools}, a mature solver based on metaheuristics, specifically crafted for routing problems.
(4) POMO \cite{kwon2020pomo}, which is recognized as the presently most potent deep model specifically designed for the direct resolution of TSP and CVRP. Here we train a POMO model on instances with three contexts simultaneously. To demonstrate the effect of lifelong learning, we additionally train the POMO incrementally on three tasks without any techniques to alleviate the catastrophic forgetting, which is denoted as POMO (lifelong) in this paper.
(5) AMDKD \cite{bi2022learning}, a generalizable neural solver for managing VRPs on different distributions using knowledge distillation. Following their settings,
we pre-train three POMO teacher models based on three distance metrics to yield a generalist student model. 
(6) Omni-VRP \cite{zhou2023towards}, a generic meta-learning framework which considers generalization across sizes and distributions in VRPs. For a fair comparison, we adapt it to our setting by sampling distinct distance metrics in each training iteration. 
For CVRP, we compare our methods with all the baselines except Concorde as it is not inherently designed for CVRP. Instead, we include another heuristic solver called hybrid genetic search (HGS) \cite{vidal2022hybrid}.
We use a default order of metrics (i.e., Euclidean$\rightarrow $Manhattan$\rightarrow $Chebyshev) for all lifelong learning methods. 

 \begin{figure}[!t]
    \centering
    \includegraphics[width=0.485\linewidth]{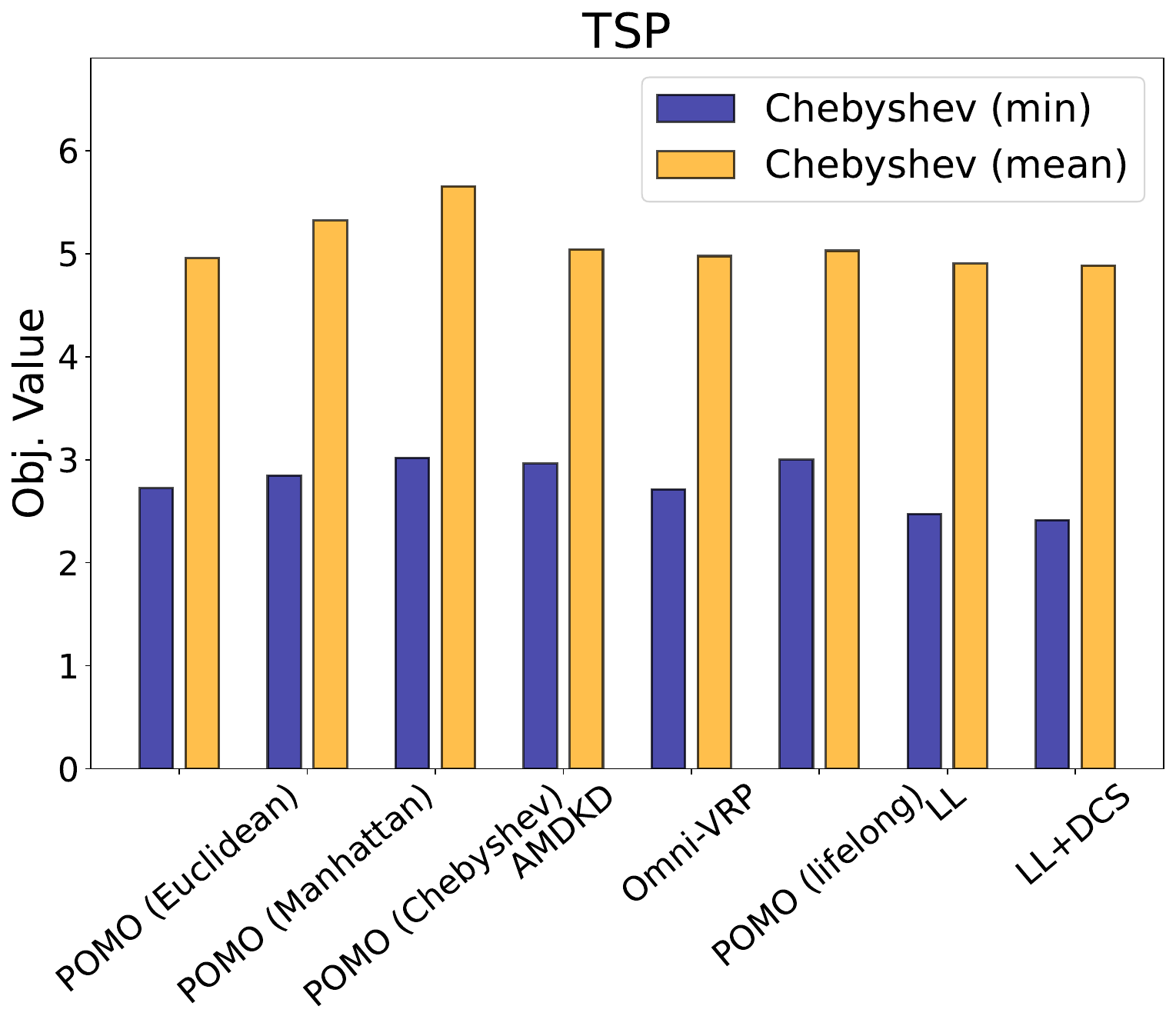}
    \includegraphics[width=0.495\linewidth]{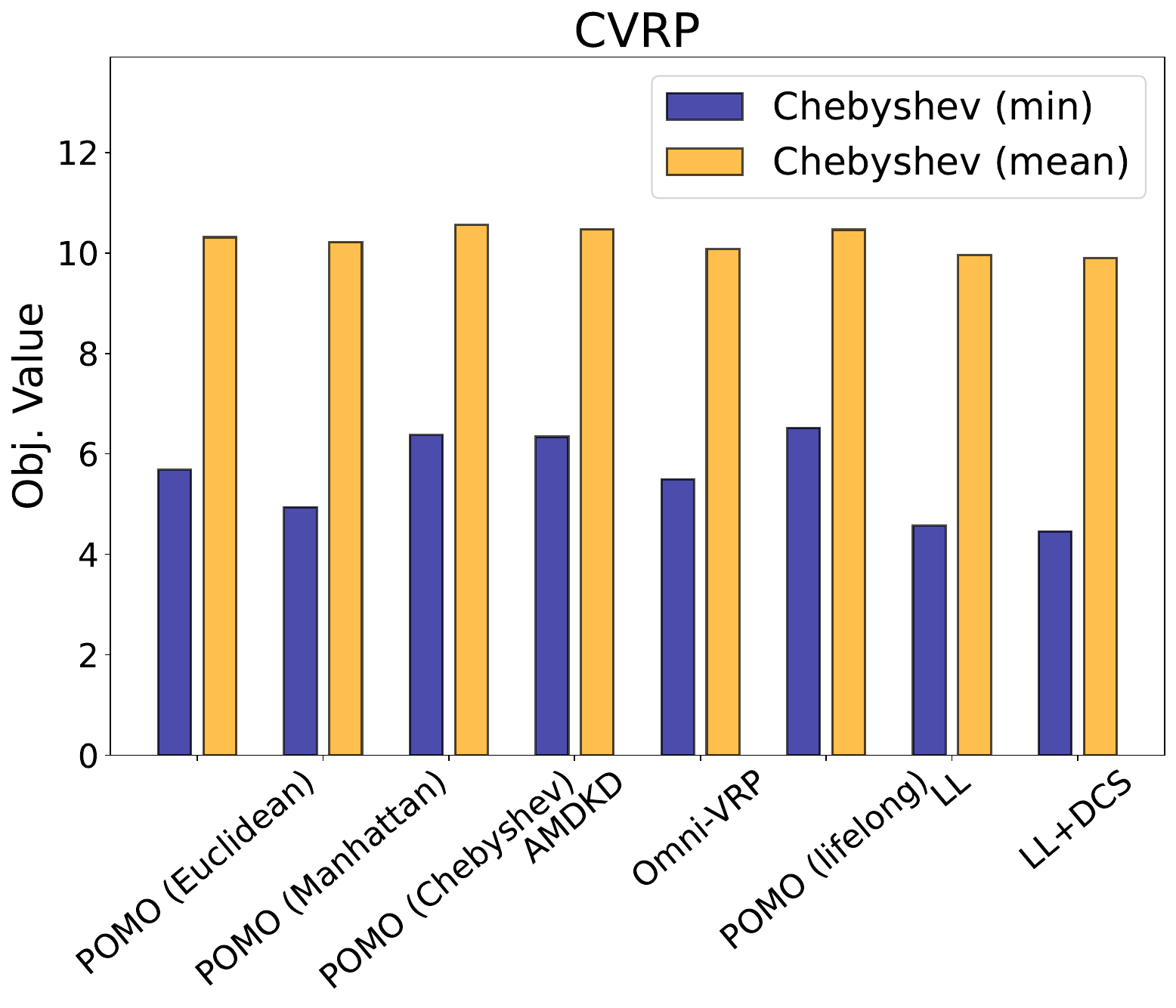}
    \vspace{-0.25in}
    \caption{Generalization on unseen metrics.}    
    \label{fig: unseen}
    \vspace{-0.15in}
\end{figure}

\input{table/cross_size2}

\subsection{Performance Comparison}
We evaluate all methods on 10,000 test instances. For each approach, we report the average objective value, average gap (relative to the best solution), and overall running time.
As shown in Table \ref{tab:performance}, our methods consistently outperform other neural solvers, demonstrating significantly lower objective values and optimality gaps. Particularly, LL and LL+DCS achieve the lowest average gaps of 1.80\% and 1.81\% on TSP, and 3.06\% and 2.69\% on CVRP, respectively.  
While AMDKD exhibits strong performance on the Chebyshev metric for TSP with a gap of 1.63\%, our LL method achieves a comparable result with a gap of 1.89\%.
Notably, LL+DCS maintains a maximum performance gap of only 3.24\% on the Chebyshev distance for CVRP.
Besides, our LL significantly surpasses OR-Tools on both TSP and CVRP. 
In particular, 
LL+DCS outperforms OR-Tools by a margin of 15.79\% (i.e., 2.32\% vs. 18.11\%) when testing on the Manhattan distance for CVRP.
Overall, both LL and LL+DCS achieve superior performance across different metrics for the two problems. 


\input{table/ablation}

\input{table/specific}

\subsection{Generalization to Unseen Metrics}\label{exp_gen}
While the effectiveness of LL has been verified on three exemplar distance metrics (i.e., Euclidean, Manhattan, and Chebyshev distance), one question is whether the proposed LL can generalize on unseen distance metrics.
To answer this question, we test all approaches on two unseen distance metrics which are deduced from the Chebyshev distance. Recall that the Chebyshev distance can be present as: $w_{\text{max}} = \max \left( |x_i - x_j|, |y_i - y_j|\right)$, where $(x_i,y_i)$ depicts the coordinate of node features.
Here we test all approaches on the following two commonly used metrics:
$w_{\text{min}} = \min \left( |x_i - x_j|, |y_i - y_j|\right)$ and $w_{\text{mean}} = \frac{1}{2} \left(|x_i - x_j|+|y_i - y_j|\right)$, which are denoted as Chebyshev (min) and Chebyshev (mean), respectively.
In addition to the previously mentioned neural models, we further present the results of three POMO models, which are trained on a specific distance metric and denoted as POMO (Euclidean), POMO (Manhattan), and POMO (Chebyshev), respectively.
As illustrated in Figure \ref{fig: unseen}, our methods significantly surpass other approaches on both TSP and CVRP, demonstrating the impressive capability in generalizing to unseen distance metrics. 
Particularly, in comparison to the highest-performing baseline models on the Chebyshev (min) metric, LL+DCS records a 10.96\% improvement over Omni-VRP on the TSP (2.414 vs. 2.711) and a 9.76\% improvement over POMO (Manhattan) on the CVRP (4.456 vs. 4.938).

\subsection{Cross-size Applications}\label{app:cross-pro}
We further investigate whether LL can effectively solve downstream VRPs in contexts of varying data sources (i.e., problem sizes). 
Specifically, we perform lifelong training for our LL and POMO (lifelong) based on the following order: 
$20\rightarrow40\rightarrow60\rightarrow80\rightarrow100\rightarrow150\rightarrow200$. 
Here our LL revisits previous problem sizes and keeps the distance metrics unchanged, as all the problems are solved with Euclidean distance.
For AMDKD, we pre-trained 7 teacher models for each training size in the range [20, 200]. Then the cross-size knowledge is distilled to a student model.  
Regarding Omni-VRP, we simply replace the distribution sampling process with size sampling during the training stage. 
Subsequently, the learned policies of all approaches are directly applied to unseen problem sizes (i.e., 300, 400, and 500) in a zero-shot manner, with their results on TSP and CVRP presented in Table \ref{cross-size}.
Overall, our methods consistently demonstrate exceptional performance across all scales of TSP and CVRP, confirming their adaptability to varying VRP sizes and strong generalization to larger, unseen problem instances.
In particular, LL+DCS shows optimal performance on all TSP instances and CVRP instances ranging from 20 to 300, while achieving comparable results to Omni-VRP on CVRP ($n$= 500). 
\input{table/sl}

\subsection{Ablation Study}\label{app:AS}
In this subsection, we conduct an ablation study to investigate the effects of key components within our framework. 
Compared with POMO (lifelong), which is trained with a lifelong learning paradigm, our LL integrates both inter-context self-attention and cross-context experience replay. Analysis of Table \ref{tab:ablation} yields the following insights: (1) Integrating inter-context self-attention (ICSA) into POMO (lifelong) significantly improves performance, reducing the average gap from 9.34\% to 2.31\% under the Manhattan distance metric. This confirms the efficacy of ICSA and its associated objective.
(2) Cross-context experience replay further boosts performance, demonstrating the effectiveness of our replay strategy and the importance of revisiting prior knowledge during lifelong learning.
In summary, we validate the efficacy of both ICSA and cross-context experience replay, underscoring their essential role in enhancing LL for solving VRPs across diverse contexts.

\input{table/lib}

\input{table/cvrplib}

\subsection{Comparison to Context-Specific Models} \label{app:spe}
In this subsection, we compare our LL methods with context-specific models on TSP.
Specifically, we present the results of metric-specific methods in Table \ref{tab:metric-specific}.
The results in Table \ref{tab:metric-specific} indicate that the metric-specific POMO models generally perform better than other counterparts when tested on their specific metric, while the proposed LL methods achieve comparable performance on both Manhattan and Chebyshev distance. Notably, our LL+DCS outperforms POMO (Euclidean) by a marginal 0.26\% when testing on the Euclidean distance on TSP. 
Overall, our LL achieves an average performance gap of 1.80\% across three metrics, surpassing all context-specific methods.
This indicates that the knowledge learned from the previous context (i.e., Euclidean distance) can be effectively preserved during the lifelong learning process, demonstrating the effectiveness of the key designs of the LL against catastrophic forgetting. 

\subsection{Versatility}
In addition to the applicability across various problem sizes, we further investigate whether our LL is versatile to different deep models.
To this end, we deploy our LL with LEHD \cite{luo2024neural} which achieves the state-of-the-art performance in solving TSP and CVRP.
We denote such a method as LL (LEHD) and perform supervised learning during the lifelong training process.
Note that the training labels are provided by LKH3 for TSP and HGS for CVRP, respectively.
Besides, we follow all the experimental settings and use the same hyperparameters as those employed for training the original LEHD. Since it is trained only with the Euclidean distance metric, therefore, we denote it as LEHD (Euclidean) in this work. Then we directly test its performance on Manhattan and Chebyshev distance metrics to demonstrate its generalization ability. 
Furthermore, we introduce another variant called LEHD (mixed metrics), that simultaneously trains LEHD using three exemplar distance metrics during the training process. 
We test the above methods with 10,000 instances and report their average objective value, average gap, and overall running time in Table \ref{tab:LEHD}.
As shown, the LL (LEHD) consistently outperforms the other two methods i.e.,  LEHD (Euclidean) and LEHD (mixed metrics) on both Manhattan distance and Chebushev distance, indicating the superior generalization ability of our lifelong learning framework. 
On the other hand, LL (LEHD) performs slightly worse than LEHD (Euclidean) on Euclidean distance (e.g., 1.97\% vs. 2.00\% for TSP). This demonstrates our LL is able to achieve comparable performance with better generalization ability for different deep learning backbones.

\subsection{Generalization to Benchmark Instances} \label{app:lib}
To further evaluate the capability of LL, we directly apply it to solve the benchmark instances from CVRPLIB~\cite{uchoa2017new} and TSPLIB~\cite{reinelt1991tsplib}. Adapted from real-world problems, 
they comprise diverse instances with unknown distributions and larger sizes.
For instance, in TSPLIB, “pcb3038” indicates a routing problem for drilling, with the size of 3,038 and edge weights representing Euclidean distances in a 2-D space. In contrast, "ali535" represents a routing problem with 535 airports around the globe and edge weights based on geographical distances.
We report the objective value of four learning-based methods on CVRP and TSP in Table \ref{CVRPLIB2}, and Table \ref{tsplib}, respectively. 
As depicted, the proposed LL consistently outperforms POMO, particularly on CVRPLIB. Similarly, Omni-VRP and AMDKD generally surpass the POMO (lifelong) method on both TSPLIB and CVRPLIB. Overall, these results affirm that the proposed LL serves as a robust neural solver for TSP and CVRP, demonstrating strong generalization capabilities, even on instances significantly different from the training set.

\section{Conclusion and Future Work}
This paper presents a lifelong learning method to solve VRPs in varying contexts. We employ a transformer network as the LL to incrementally learn constructing solutions in a series of contexts. The inter-context self-attention mechanism is proposed to transfer useful knowledge from the preceding contexts to avoid catastrophic forgetting. Furthermore, we replay the previous distance metrics to consolidate the learned policies in the preceding contexts, when training the LL in each context. The proposed method is evaluated to solve VRPs in contexts of varying distance metrics and problem sizes, respectively. Results show that the trained solvers can handle various contexts with favorable performance, meaning better versatility to cope with VRPs in more diverse scenarios. The ablation and benchmarking studies further verify the efficacy of lifelong learning and its key designs. 
In future work, we intend to apply the LL to solve VRPs in contexts with multiple varying dimensions, e.g., by changing problem sizes and objectives concurrently.


\bibliographystyle{IEEEtran}
\bibliography{ref}
\end{document}

%% file: table/performance.tex
\begin{table*}[!ht] 
\vspace{-0.1in}
  \caption{Results on three different metrics with size 100.}
  \label{tab:performance}
\scriptsize
  \vspace{-0.1in}
  \begin{center}
  \renewcommand\arraystretch{0.9}
  \begin{tabular}{cl|ccc|ccc|ccc|c}
    \toprule
    \midrule
    &\multirow{2}{*}{Methods}&\multicolumn{3}{c|}{Euclidean}&\multicolumn{3}{c|}{Manhattan}&\multicolumn{3}{c|}{Chebyshev}& \multirow{2}{*}{Avg. Gap} \\
    &&\multicolumn{1}{c}{Obj.} & \multicolumn{1}{c}{Gap} &\multicolumn{1}{c|}{Time}
    & \multicolumn{1}{c}{Obj.} & \multicolumn{1}{c}{Gap}& \multicolumn{1}{c|}{Time}
    &\multicolumn{1}{c}{Obj.}& \multicolumn{1}{c}{Gap}&\multicolumn{1}{c|}{Time} & \\

     \cmidrule(lr){1-12}
     \multirow{9}{*}{\rotatebox{90}{TSP}}
     &Concorde  & 7.708&0.00\%&1h &9.684&0.01\%&1.28h &6.919&0.00\%&1.02h & 0.00\% \\
     &LKH3            &7.762&0.70\%&50m  &9.677&0.00\%&56m &6.920&0.01\%&54m & 0.24\% \\
     &OR-Tools        &8.011 &3.93\%&57m &10.119&4.57\%&56m &7.203&4.10\%&57m & 4.20\% \\ 
     \cmidrule(lr){2-12}
     &POMO & 7.916  &2.70\%&11s& 9.984 &3.17\%&11s& 7.131&3.06\%&11s & 2.98\% \\ 
     &POMO (lifelong)  & 7.923 &2.79\%&11s& 9.877&2.07\%&11s& 7.205&4.13\%&11s & 3.00\% \\ 
     &AMDKD          & 7.898 &2.46\%&11s& 10.083&4.20\%& 11s& $\textbf{7.034}$ &$\textbf{1.63\%}$ &11s & 2.76\% \\
     &Omni-VRP          &7.860&1.97\%&11s&9.913&2.44\%&11s&7.086&2.41\%&11s & 2.27\% \\     
     &LL (Ours) & 7.828 & 1.56\% &11s & $\textbf{9.872}$ & $\textbf{1.94\%}$ &11s & 7.050&1.89\%&11s & \textbf{1.80\%} \\ 
     &LL+DCS (Ours) & $\textbf{7.801}$ & $\textbf{1.21\%}$ &11s & 9.898 & 2.28\% &11s & 7.054&1.95\%&11s & 1.81\% \\ 
    \midrule

     \multirow{9}{*}{\rotatebox{90}{CVRP}}    
     &HGS              &15.538&0.00\%&38m&19.303&0.00\%&38m&13.716&0.00\%&38m & 0.00\% \\
     &LKH3             &15.680&0.91\%&1.33h  &19.452&0.77\%&1.42h &13.810&0.65\%&1.27h & 0.78\% \\
     &OR-Tools         &18.043&16.06\%&2.8h &22.799&18.11\%&2.8h&16.035&16.91\%&2.8h & 17.03\% \\  
     \cmidrule(lr){2-12}  
     &POMO & 16.376  &5.39\%&13s& 20.223 &4.77\%&13s& 14.477&5.55\%&13s & 5.24\% \\ 
     &POMO (lifelong)  & 16.416 &5.65\%&13s& 21.105&9.34\%&13s & 14.468&5.48\%&13s & 6.82\% \\ 
     &AMDKD            &16.228 &4.44\%&13s& 20.394&5.65\%&13s & 14.423&5.15\%&13s & 5.08\% \\
     &Omni-VRP         &15.943 &2.61\%&13s&20.154&4.41\%&13s &14.311&4.34\%&13s & 3.79\% \\     
     &LL (Ours)& $\textbf{15.905}$ & $\textbf{2.36}\%$ &13s & 19.852 & 2.84\% &13s &14.262&3.98\%&13s & 3.06\% \\ 
     &LL+DCS (Ours)& 15.929 & 2.51\% &13s & $\textbf{19.751}$ & $\textbf{2.32\%}$&13s &$\textbf{14.161}$&$\textbf{3.24\%}$&13s & \textbf{2.69\%} \\ 
    \midrule
    \bottomrule
  \end{tabular}
  \end{center}
 \vspace{-0.15in}
\end{table*}

%% file: table/cross_size2.tex
\begin{table*}[!t] 
\centering
\scriptsize
\vspace{-0.1in}
  \caption{Cross-size generalization on TSP and CVRP.}
  \label{cross-size}
  \vspace{-0.1in}
  \begin{center}
  \renewcommand\arraystretch{0.7}
  \begin{tabular}{cl|cc|cc|cc|cc|cc|cc}
    \toprule
    \midrule
    &\multirow{2}{*}{Methods}&\multicolumn{2}{c|}{20}&\multicolumn{2}{c|}{60}&\multicolumn{2}{c|}{100}&\multicolumn{2}{c|}{200}&\multicolumn{2}{c|}{300}&\multicolumn{2}{c}{500}
    \\
    
    &&\multicolumn{1}{c}{Obj.} & \multicolumn{1}{c|}{Gap} 
    &\multicolumn{1}{c}{Obj.} & \multicolumn{1}{c|}{Gap} 
    &\multicolumn{1}{c}{Obj.} & \multicolumn{1}{c|}{Gap} 
    &\multicolumn{1}{c}{Obj.} & \multicolumn{1}{c|}{Gap} 
    &\multicolumn{1}{c}{Obj.} & \multicolumn{1}{c|}{Gap} 
    &\multicolumn{1}{c}{Obj.} & \multicolumn{1}{c}{Gap} \\
    
    \midrule
     \multirow{9}{*}{\rotatebox{90}{TSP}}&Concorde &3.830 &0.00\% &6.167 &0.00\% &7.708 &0.00\% &10.708 &0.14\% &12.934 &0.02\% &16.521 &0.01\% \\
    & LKH3 & 3.831 &0.03\% &6.168 &0.02\% &7.759 &0.66\% &10.693 &0.00\% & 12.932 &0.00\% & 16.520 &0.00\% \\     
     &OR-Tools & 3.874 &1.15\% & 6.379 &3.43\% &8.011 &3.93\% &11.063 &3.46\% & 13.410 &3.70\% & 17.128 &3.68\% \\
           \cmidrule(lr){2-14}
     
     &POMO (lifelong)   & 3.958 &3.34\% & 6.288 &1.96\% & 7.864 &2.02\% & 10.855 &1.52\% & 13.281 &2.70\% & 17.873 &8.19\% \\
     &AMDKD &3.888 &1.51\% &6.385 &3.54\% &8.255 &7.10\% &12.231 &14.38\% & 15.677 &21.23\% & 21.651 &31.06\% \\ 
     &Omni-VRP & 3.897 &1.75\% &6.302 &2.19\% &7.933 &2.92\% &10.992 &2.80\% & 13.485 &4.28\% & 17.943 &8.61\% \\      
    
     &LL& 3.833 &0.08\% &6.188 &0.34\% &$\textbf{7.810}$ &$\textbf{1.32\%}$ &10.850 &1.47\% & 13.273 &2.64\% & 17.741 &7.38\% \\

     &LL+DCS & $\textbf{3.832}$ & $\textbf{0.05\%}$ & $\textbf{6.184}$ & $\textbf{0.28\%}$ & 7.812 & 1.35\% & $\textbf{10.822}$ & $\textbf{1.21\%}$ & $\textbf{13.241}$ & $\textbf{2.39\%}$ & $\textbf{17.642}$ & $\textbf{6.80\%}$ \\

    \midrule
          \multirow{8}{*}{\rotatebox{90}{CVRP}} &HGS &6.017 &0.00\% &11.762 &0.00\% &15.928 &0.00\% &21.727 &0.00\% &26.091 &0.00\% &32.617 &0.00\% \\
     &LKH3 & 6.050 &0.55\% & 11.811 &0.42\% &15.962 &0.21\% & 22.068 &1.62\% & 26.150 &0.23\% & 32.697 &0.24\% \\     

           \cmidrule(lr){2-14}
     &POMO (lifelong)   & 8.583 &42.68\% & 12.651 &7.56\% & 16.208 &1.76\% & 22.423 &3.20\% & 26.940 &3.25\% & 39.210 &20.20\% \\
     &AMDKD &7.684 &27.72\% &12.138 &3.20\% &16.231 &1.90\% &23.597 &8.61\% & 29.123 &11.62\% & 38.155 &16.97\% \\ 
     &Omni-VRP & 7.134 &18.55\% &12.149 &3.29\% &16.125 &1.24\% &22.701 &4.48\% & 27.001 & 3.49\% & $\textbf{33.077}$ & $\textbf{1.41\%}$ \\      

     &LL & 6.268 &4.17\% & 11.936 &1.48\% & 16.055 &0.80\% &22.466 &3.40\% & 26.889 &3.06\% & 33.642 &3.14\% \\

     & LL+DCS & $\textbf{6.251}$ & $\textbf{3.89\%}$ & $\textbf{11.870}$ & $\textbf{0.92\%}$ & $\textbf{15.948}$ & $\textbf{0.13\%}$ & $\textbf{22.414}$ & $\textbf{3.16\%}$ & $\textbf{26.857}$ & $\textbf{2.94\%}$ & 33.479 &2.64\% 
 \\   
    \midrule
    \bottomrule
  \end{tabular}
  
  \end{center}
  \vspace{-0.25in}
\end{table*}

%% file: table/ablation.tex
\begin{table}[!tp] 
  \caption{Ablation study on CVRP with problem size 100. ICSA denotes the inter-context self-attention.}
  \label{tab:ablation}
  \vspace{-0.2in}
  \scriptsize
  \begin{center}
  \renewcommand\arraystretch{0.9}
 \resizebox{0.49\textwidth}{!}{ 
  \begin{tabular}{l|ccc|ccc|ccc}
    \toprule
    \midrule

    \multirow{2}{*}{Methods}&\multicolumn{3}{c|}{Euclidean}&\multicolumn{3}{c|}{Manhattan}&\multicolumn{3}{c}{Chebyshev}
    \\
    &\multicolumn{1}{c}{Obj.} & \multicolumn{1}{c}{Gap} &\multicolumn{1}{c|}{Time}
    & \multicolumn{1}{c}{Obj.} & \multicolumn{1}{c}{Gap}& \multicolumn{1}{c|}{Time}
    &\multicolumn{1}{c}{Obj.}& \multicolumn{1}{c}{Gap}&\multicolumn{1}{c}{Time} \\

     \cmidrule(lr){1-10}

     HGS              &15.538&0.00\%&38min&19.303&0.00\%&38min&13.716&0.00\%&38min \\
     LKH3             &15.680&0.91\%&1.33h&19.452&0.71\%&1.42h&13.805&0.65\%&1.27h \\
      \cmidrule(lr){1-10}
     POMO (lifelong)  & 16.416 &5.65\% &13s& 21.105 &9.34\%&13s & 14.468&5.48\%&13s \\ 
     + ICSA &16.077&3.47\% &13s&19.902&2.31\%&13s & 14.430&5.20\%&13s\\
     LL (Ours) & $\textbf{15.905}$ &$\textbf{2.36\%}$ &13s & 19.852&  2.84\%&13s &14.262& 3.98\%&13s \\ 
     LL+DCS (Ours)& 15.929 & 2.51\% &13s & $\textbf{19.751}$ & $\textbf{2.32\%}$&13s &$\textbf{14.161}$&$ \textbf{3.24\%}$ &13s \\ 
    \midrule
    \bottomrule
   
    \end{tabular}}
  \end{center}
   \vspace{-0.2in}
\end{table}

%% file: table/specific.tex
\begin{table}[!tp] 
 \caption{Comparison with three context-specific POMO models on TSP with problem size 100.} 
 \vspace{-0.2in}
  \label{tab:metric-specific}
  \scriptsize
  \begin{center}
 \resizebox{0.49\textwidth}{!}{ 
  \begin{tabular}{l|ccc|ccc|ccc|c}
    \toprule
    \midrule

    \multirow{2}{*}{Methods}&\multicolumn{3}{c|}{Euclidean}&\multicolumn{3}{c|}{Manhattan}&\multicolumn{3}{c}{Chebyshev} &\multirow{2}{*}{Ave Gap}
    \\
    &\multicolumn{1}{c}{Obj.} & \multicolumn{1}{c}{Gap} &\multicolumn{1}{c|}{Time}
    & \multicolumn{1}{c}{Obj.} & \multicolumn{1}{c}{Gap}& \multicolumn{1}{c|}{Time}
    &\multicolumn{1}{c}{Obj.}& \multicolumn{1}{c}{Gap}&\multicolumn{1}{c}{Time} \\

     \cmidrule(lr){1-11}
     POMO (Euclidean)  & \underline{7.821} &\underline{1.47\%} &11s& 9.912&2.43\%&11s& 7.071&2.20\%&11s & \underline{2.03\%} \\
     POMO (Manhattan)  & 7.916 &2.70\%&11s  & \underline{9.853}&\underline{1.82\%}&11s& 7.226&4.44\%&11s &2.99\%\\
     POMO (Chebyshev)  & 7.907 &2.58\%&11s& 10.121&4.59\%&11s& \underline{7.033}&\underline{1.65\%}&11s & 2.94\%\\

    \cmidrule(lr){1-11}
    POMO (lifelong)  & 7.923 &2.79\%&11s& 9.877&2.07\%&11s& 7.205&4.13\%&11s &3.00\%\\ 
    LL (Ours) & 7.828 & 1.56\% &11s & \textbf{9.872} & \textbf{1.94\%} &11s & $\textbf{7.050}$&$\textbf{1.89\%}$&11s &\textbf{1.80\%} \\ 
    LL+DCS (Ours) & $\textbf{7.801}$ & $\textbf{1.21\%}$ &11s & 9.898 & 2.28\% &11s & 7.054&1.95\%&11s &1.81\% \\ 
    \midrule
    \bottomrule
 \end{tabular}}
  \end{center}
  \vspace{-0.2in}
\end{table}

%% file: table/sl.tex
\begin{table}[!t] 
  \caption{Results on three different metrics with size 100.  We deploy LL with LEHD to demonstrate its versatility.}
  \label{tab:LEHD}
  \vspace{-0.2in}
  \scriptsize
  \begin{center}
  \renewcommand\arraystretch{1.1}
  \resizebox{0.49\textwidth}{!}{ 
  \begin{tabular}{cl|ccc|ccc|ccc}
    \toprule
    \midrule

    &\multirow{2}{*}{Methods}&\multicolumn{3}{c|}{Euclidean}&\multicolumn{3}{c|}{Manhattan}&\multicolumn{3}{c}{Chebyshev}
    \\
    &&\multicolumn{1}{c}{Obj.} & \multicolumn{1}{c}{Gap} &\multicolumn{1}{c|}{Time}
    & \multicolumn{1}{c}{Obj.} & \multicolumn{1}{c}{Gap}& \multicolumn{1}{c|}{Time}
    &\multicolumn{1}{c}{Obj.}& \multicolumn{1}{c}{Gap}&\multicolumn{1}{c}{Time} \\

     \cmidrule(lr){1-11}
     \multirow{7}{*}{\rotatebox{90}{TSP}}
          &Concorde  & 7.708&0.00\%&1h &9.684&0.01\%&1.28h &6.919&0.00\%&1.02h \\
     &LKH3            &7.762&0.70\%&50m  &9.677&0.00\%&56m &6.920&0.01\%&54m \\
     &OR-Tools        &8.011 &3.93\%&57m &10.119&4.57\%&56m &7.203&4.10\%&57m \\ \cmidrule(lr){2-11}
     &LEHD (Euclidean)&  $\textbf{7.860}$ & $\textbf{1.97\%}$ &11s & 9.963 & 2.96\% &11s & 7.088 &2.44\%&11s \\
     &LEHD (mixed metrics)&  7.873 & 2.14\% &11s & 9.928 & 2.59\% &11s & 7.086 &2.41\%&11s \\
     &LL (LEHD)&  7.862 & 2.00\% &11s & $\textbf{9.906}$ & $\textbf{2.37\%}$ &11s & $\textbf{7.083}$ &$\textbf{2.37\%}$&11s \\
    \midrule

     \multirow{7}{*}{\rotatebox{90}{CVRP}}    
&HGS              &15.538&0.00\%&38m&19.303&0.00\%&38m&13.716&0.00\%&38m \\
     &LKH3             &15.680&0.91\%&1.33h  &19.452&0.77\%&1.42h &13.810&0.65\%&1.27h \\
     &OR-Tools         &18.043&16.06\%&2.8h &22.799&18.11\%&2.8h&16.035&16.91\%&2.8h \\
     \cmidrule(lr){2-11}
    &LEHD (Euclidean)&  $\textbf{16.393}$ & $\textbf{5.50\%}$ &13s & 21.241 & 10.00\% &13s & 14.826 &8.10\%&13s \\
     &LEHD (mixed metrics)&  16.621 & 7.00\% &13s & 21.171 & 9.68\% &13s & 14.785 &7.80\%&13s \\
     &LL (LEHD)&  16.545 & 6.48\% &13s & $\textbf{21.141}$ & $\textbf{9.52\%}$ &13s & $\textbf{14.676}$ &$\textbf{7.00\%}$ &13s \\
    \midrule
    \bottomrule
  \end{tabular}}
  \end{center}
  \vspace{-0.2in}
\end{table}

%% file: table/lib.tex
\begin{table*}[!htp]
    \centering
    \scriptsize
    \caption{Results on TSPlib instances. Problem size from 51 to 18,512.}
    \label{tsplib}
    \renewcommand\arraystretch{1.25}
    \setlength{\tabcolsep}{3.5pt}
     \resizebox{0.8\textwidth}{!}{
    \begin{tabular}{l|ccccc|l|ccccc}
        \toprule
        \midrule
        Instances & LL+DCS & LL & AMDKD & POMO (lifelong) & Omni-VRP & Instances & LL+DCS & LL & AMDKD & POMO (lifelong) & Omni-VRP \\
        \midrule
        
a280&2929&\textbf{2905}&2924&3014&2992&bier127&\textbf{122638}&124791&125902&127774&142444\\
brd14051&886124&897180&\textbf{870883}&971460&900805&d1291&86535&76706&85991&\textbf{76054}&82156\\
d15112&2810579&\textbf{2765946}&2936123&3090090&2888761&d1655&106522&\textbf{91892}&104437&96197&108364\\
d18512&1192088&\textbf{1186789}&1205287&1294978&1253215&d198&20918&21255&\textbf{19667}&20052&24484\\
d2103&121908&\textbf{117984}&153065&126632&123997&d493&63260&78613&\textbf{56094}&57763&60388\\
d657&70711&62705&80235&\textbf{61410}&69568&eil101&651&675&\textbf{645}&662&654\\
eil51&\textbf{449}&470&485&449&451&eil76&559&562&564&569&\textbf{554}\\
fl1400&31541&30639&\textbf{27204}&28005&30844&fl1577&33200&32983&35061&34383&\textbf{32958}\\
fl3795&51854&51574&53501&53389&\textbf{48368}&fl417&14827&15353&\textbf{13712}&15218&13834\\
fnl4461&\textbf{282312}&286105&288733&308508&292052&gil262&\textbf{2396}&2478&2477&2464&2507\\
kroA100&\textbf{21915}&22703&21994&22316&22046&kroA150&\textbf{27170}&28529&27820&27656&27466\\
kroA200&\textbf{29923}&30804&30604&30770&30981&kroB100&22605&23074&22755&\textbf{22574}&23183\\
kroB150&\textbf{26168}&27567&27048&27177&27342&kroB200&\textbf{30721}&30915&31791&31607&32373\\
kroC100&\textbf{20787}&21366&21116&21273&21889&kroD100&\textbf{21644}&22145&21896&22061&21965\\
kroE100&\textbf{22284}&22559&22615&22694&22955&nrw1379&\textbf{76631}&78816&80284&81546&78791\\
p654&45408&42758&47361&\textbf{42687}&45644&pcb1173&76149&\textbf{75762}&78786&82931&76384\\
pcb3038&202082&\textbf{200383}&213949&225963&204053&pcb442&\textbf{56834}&57805&59639&59004&58446\\
pr1002&343088&354571&356359&354405&\textbf{341564}&pr107&51599&52490&\textbf{49931}&50257&50585\\
pr124&\textbf{61009}&62522&61301&62002&61687&pr136&101146&101081&\textbf{100412}&101280&102430\\
pr144&\textbf{60756}&62111&61474&61333&62871&pr152&\textbf{74419}&76212&77887&77148&77723\\
pr226&\textbf{82488}&83754&84351&84688&85532&pr2392&\textbf{553169}&562316&583754&598908&561455\\
pr264&\textbf{60104}&61303&68011&62024&60433&pr299&\textbf{54287}&56496&57569&56252&57783\\
pr439&\textbf{126640}&131062&132476&137961&132196&pr76&109414&109464&109830&\textbf{109254}&111592\\
rat195&2628&2703&\textbf{2617}&2666&2722&rat575&8263&\textbf{8177}&8408&8182&8248\\
rat783&\textbf{11064}&11105&11364&11565&11099&rat99&\textbf{1270}&1317&1293&1285&1349\\
rd100&7951&8183&\textbf{7945}&7987&7963&rd400&\textbf{16926}&17199&17449&17332&17235\\
rl11849&1707060&\textbf{1699835}&1750588&1994249&1716573&rl1304&352516&\textbf{341538}&363073&390256&345541\\
rl1323&386315&382407&400623&407946&\textbf{374653}&rl1889&468564&466680&489740&489999&\textbf{463753}\\
rl5915&\textbf{954809}&992798&1039307&1103535&998196&rl5934&973688&\textbf{952043}&1034868&1082503&988151\\
st70&702&702&\textbf{700}&703&706&ts225&136509&140661&\textbf{131990}&142152&141435\\
tsp225&\textbf{4251}&4331&4443&4441&4350&u1060&305996&307163&321035&330505&\textbf{304943}\\
u1432&\textbf{191767}&198530&199891&206377&197675&u159&44789&46006&\textbf{44708}&45027&46262\\
u1817&84387&\textbf{83985}&89792&88553&86645&u2152&\textbf{94462}&97428&100313&102251&98871\\
u2319&\textbf{281164}&283737&286770&301983&291946&u574&45217&46281&46333&48328&\textbf{44901}\\
u724&52238&\textbf{52057}&54007&54207&53072&vm1084&314342&317614&325405&324163&\textbf{303988}\\
        \midrule
        \bottomrule
    \end{tabular}}
\end{table*}

%% file: table/cvrplib.tex
\begin{table*}[!ht] 
\centering
\tiny
  \caption{Results on CVRPlib (Set X) instances. Problem size from 101 to 916.}
  \label{CVRPLIB2}
    \renewcommand\arraystretch{1.0}
        \setlength{\tabcolsep}{3pt}
    \resizebox{0.8\textwidth}{!}{ 
    \begin{tabular}{@{}l|ccccc|l|ccccc@{}}
    \toprule
    \midrule
    Instances & LL+DCS & LL & AMDKD & POMO (lifelong) & Omni-VRP & Instances & LL+DCS & LL & AMDKD & POMO (lifelong) & Omni-VRP \\
    \midrule

X-n101-k25&\textbf{29074}&29562&30050&29702&29606&X-n106-k14&\textbf{27400}&27504&27998&27530&61018\\
X-n120-k6&\textbf{14286}&15075&14604&14442&15097&X-n125-k30&59425&59427&\textbf{58556}&59264&71844\\
X-n129-k18&29615&\textbf{29433}&29985&30515&66727&X-n134-k13&\textbf{11457}&11594&11674&11669&43435\\
X-n139-k10&\textbf{13865}&14242&14205&14088&13927&X-n143-k7&\textbf{16165}&16847&17519&17005&18428\\
X-n148-k46&\textbf{46433}&56515&48407&47099&47028&X-n153-k22&\textbf{23410}&23883&24287&23952&24151\\
X-n176-k26&\textbf{52340}&52433&54946&53693&116909&X-n181-k23&\textbf{25989}&26279&43172&26850&27657\\
X-n186-k15&\textbf{25560}&25629&25881&25561&60732&X-n190-k8&20478&19789&\textbf{18900}&19069&44672\\
X-n204-k19&\textbf{20654}&20965&20782&21431&21087&X-n209-k16&32558&\textbf{32345}&32721&32693&35184\\
X-n214-k11&12353&\textbf{11997}&12185&12317&12898&X-n219-k73&134781&129484&124472&\textbf{121837}&180367\\
X-n223-k34&42651&\textbf{42337}&43688&44128&43785&X-n228-k23&\textbf{28259}&28452&29828&29558&29179\\
X-n233-k16&\textbf{20681}&21090&20837&21499&20804&X-n237-k14&30264&29940&32408&\textbf{29401}&30634\\
X-n242-k48&\textbf{86788}&89048&88489&88345&158694&X-n247-k50&43210&42809&\textbf{41922}&43245&62550\\
X-n251-k28&40806&\textbf{40465}&40957&41311&95860&X-n256-k16&20646&22272&20819&20906&\textbf{20488}\\
X-n270-k35&\textbf{37081}&37239&37805&38469&38547&X-n275-k28&22832&\textbf{22542}&22714&22894&23066\\
X-n280-k17&\textbf{36195}&36207&37465&36846&123292&X-n284-k15&\textbf{22010}&23573&22776&22934&54525\\
X-n351-k40&29166&\textbf{28416}&29231&30431&30270&X-n359-k29&55244&\textbf{54760}&55781&55894&173858\\
X-n367-k17&25916&\textbf{25234}&26474&27219&27883&X-n376-k94&177378&194089&166565&\textbf{157581}&253768\\
X-n401-k29&70847&\textbf{70135}&79613&71695&168775&X-n411-k19&23356&\textbf{23135}&24506&23849&24393\\
X-n439-k37&40864&41724&\textbf{39920}&41854&40864&X-n449-k29&60236&\textbf{60211}&60805&62315&224478\\
X-n459-k26&29913&30740&\textbf{27838}&28917&29434&X-n469-k138&259343&287197&\textbf{249686}&250361&353097\\
X-n480-k70&\textbf{96720}&121540&98058&102154&101138&X-n491-k59&74236&\textbf{73581}&74752&76513&75747\\
X-n573-k30&82677&75833&67671&\textbf{57267}&160214&X-n586-k159&230463&242477&\textbf{214775}&226573&321275\\
X-n627-k43&74607&73695&\textbf{71583}&75700&142183&X-n641-k35&76822&\textbf{73065}&96319&73620&118132\\
X-n655-k131&134262&146973&129849&\textbf{120473}&132631&X-n670-k130&\textbf{168591}&177384&172826&175814&304822\\
X-n685-k75&81207&\textbf{78395}&79183&83850&83842&X-n701-k44&92327&\textbf{91125}&91383&92329&270431\\
X-n749-k98&90592&88877&\textbf{87821}&93509&92253&X-n766-k71&128361&\textbf{128152}&132252&131622&392648\\
X-n783-k48&87645&87892&\textbf{83550}&91338&129436&X-n801-k40&108972&88254&159116&\textbf{88058}&126118\\
X-n856-k95&105205&142277&107275&107791&\textbf{101706}&X-n876-k59&116540&\textbf{114344}&270266&116301&455275\\
X-n895-k37&73628&123171&\textbf{65493}&72225&72201&X-n916-k207&\textbf{368235}&374550&393190&375595&648502\\
    \midrule
    \bottomrule
\end{tabular}}
\vspace{0.1in}
\end{table*}